\documentclass{article}

\PassOptionsToPackage{round}{natbib}

\usepackage[preprint]{neurips_2024}

\usepackage[utf8]{inputenc} 
\usepackage[T1]{fontenc}    
\usepackage{hyperref}       
\usepackage{url}            
\usepackage{booktabs}       
\usepackage{amsfonts}       
\usepackage{nicefrac}       
\usepackage{microtype}      
\usepackage{xcolor}         
\usepackage{graphicx}
\usepackage{amsmath}
\usepackage{algorithm}
\usepackage{algpseudocode}
\usepackage{booktabs}
\usepackage{multirow} 

\title{Text Guided Image Editing with Automatic Concept Locating and Forgetting}

\author{Jia Li\textsuperscript{1,2,3}, Lijie Hu\textsuperscript{1,2}, Zhixian He\textsuperscript{6}, Jingfeng Zhang\textsuperscript{4}, Tianhang Zheng\textsuperscript{5}, Di Wang\textsuperscript{1,2} \\
\textsuperscript{1} Provable Responsible AI and Data Analytics (PRADA) Lab \\ \textsuperscript{2} King Abdullah University of Science and Technology \\ \textsuperscript{3} Chinese Academy of Sciences \quad \textsuperscript{4} University of Auckland \\ \textsuperscript{5} University of Missouri \quad\textsuperscript{6}Sun Yat-sen University}

\begin{document}

\maketitle

\begin{abstract}
\label{main-abs}

With the advancement of image-to-image diffusion models guided by text, significant progress has been made in image editing. However, a persistent challenge remains in seamlessly incorporating objects into images based on textual instructions, without relying on extra user-provided guidance. Text and images are inherently distinct modalities, bringing out difficulties in fully capturing the semantic intent conveyed through language and accurately translating that into the desired visual modifications. Therefore, text-guided image editing models often produce generations with residual object attributes that do not fully align with human expectations. To address this challenge, the models should comprehend the image content effectively away from a disconnect between the provided textual editing prompts and the actual modifications made to the image. In our paper, we propose a novel method called Locate and Forget (LaF), which effectively locates potential target concepts in the image for modification by comparing the syntactic trees of the target prompt and scene descriptions in the input image, intending to forget their existence clues in the generated image. Compared to the baselines, our method demonstrates its superiority in text-guided image editing tasks both qualitatively and quantitatively.

\end{abstract}

\section{Introduction}
\label{main-intro}

\begin{figure}[h]
    \centering
    \includegraphics[width=0.9\textwidth]{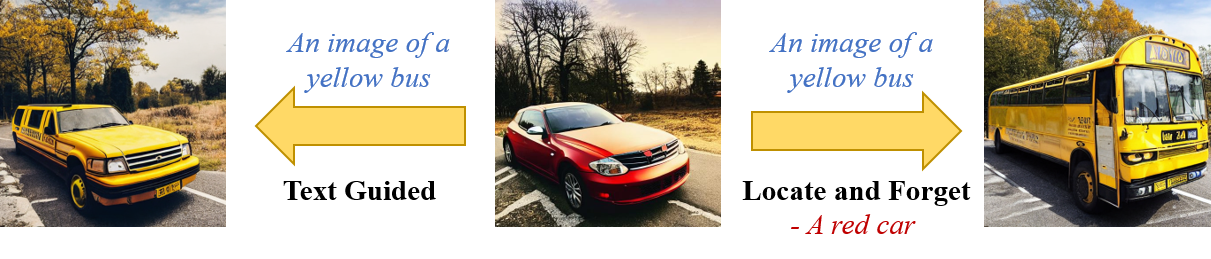}
    \caption{Original image presents a red car. When the input text instruction is \textit{an image of a yellow bus}, Stable Diffusion focuses on modifying the color but preserves the old shape. By analyzing the scene description of the image, concepts that users intend to edit are located and forgotten in the denoising steps for an improved output.}
    \label{fig:bg}
\end{figure}


Diffusion models are widely applied to image editing tasks \citep{ddim,ddpm,ldm}. They are particularly effective for tasks such as image denoising, inpainting, super-resolution and style transfer \citep{avrahami2022blended,kawar2023imagic,nichol2021glide,yu2023inpaint}. Text conditions are a common and versatile form of instruction. By using text conditions, users can guide the image editing process through simple textual descriptions without requiring expertise in complex editing tools or techniques \citep{diffusion-c-free}.

While these methods show promising results in image editing, they always struggle to synthesize the image information and targeted editing tasks due to modality gaps between images and limited textual information \citep{metri_clip_score,blip}. Text conditions often lack the full context of the image, making it difficult to accurately identify the core conceptual elements within the image that need to be transformed. As a result, the edits made may contradict the user's intended objectives and expectations. For example, as shown in Figure \ref{fig:bg}, existing text-guided image editing models can become confused about the specific concept, like shape and color, of the image that needs to be modified just based on the provided textual instructions.

A fundamental challenge in text-guided image editing from the semantic gap that often exists between the natural language used in the textual conditions and the visual representations in the image. The rich, contextual meaning conveyed through language does not always translate seamlessly to the pixels and visual elements that make up an image, leading to disconnects between the textual directives and the resulting transformations \citep{song2023bridge,liu2023visual,vit}. To overcome these issues, it is common to introduce some additional form of supervision, which can be implemented in various ways. One approach is to provide additional annotation information, such as manually marking the regions to be modified \citep{avrahami2022blended,yu2022scaling,rombach2022high,hertz2022prompt,patashnik2023localizing}. This allows the model to more accurately locate and modify the target regions with user-customized instruction, although relying on such manual annotations can be labour-intensive and limit the scalability of text-guided image editing systems.

In this paper, Our method proposes a new method named Locate and Forget (LaF) to automatically assist the diffusion models to locate where is intended to be edited and selectively forget the concept of removal. The scene description, generated through image content, provides a structured textual representation of the objects, attributes, and relationships within the image, where the model can learn to understand the contextual relationships between different concepts \citep{scenegraph1,scenegraph2,vlm}. By comparing this scene description with the input text prompt, the LaF model can infer the user's editing intent and selectively target the relevant visual elements for transformation, while avoiding unwanted changes to the unrelated parts of the image. The forgetting process of the irrelevant concepts allows the model to focus its efforts on generating outputs that closely align with the preferences and concepts expressed in the textual instructions \citep{diffusion-c-free}. By precisely locating and modifying the appropriate elements, LaF represents a significant advancement in the field of text-guided image editing:

\begin{enumerate}
    \item We propose an efficient framework for image editing in text-guided diffusion models, which releases the burden on the user to provide explicit instructions for every edit, enhancing the convenience and efficiency of the editing process.
    \item Our method introduces the scene description into image editing, providing provides additional visual information that helps the model better understand the relationships between scenes and objects within the image. It effectively leverages the structured information present in the image, enhancing the controllability and effectiveness of image editing tasks.
    \item Our method exhibits qualitative and quantitative improvements in text-guided image editing tasks when compared to the baselines.
\end{enumerate}

\section{Related Works}

\textbf{Image editing in diffusion models.}
In recent years, Diffusion Models (DMs) have emerged as the new de facto standard in the field of image generation \citep{ho2020denoising, nichol2021glide, podell2023sdxl, hu2024anomalydiffusion}. However, adapting them for image editing tasks remains challenging. This primarily arises from an inherent conflict between the original objective of Diffusion Models to reconstruct images "from scratch" while the goal of generating editable representations that adhere to specific trained distributions \citep{huang2024diffusion}. Several works have already made valuable attempts to address this issue. For instance, some scholars have utilized masks to delineate target editing regions \citep{avrahami2022blended, yu2023inpaint, rombach2022high, li2023gligen}, while others focused on manipulating attention weights during the diffusion process, thereby directing the model's focus towards specific areas of the image \citep{hertz2022prompt, Park_2024_WACV, patashnik2023localizing, liu2024understanding}. Furthermore, some studies have demonstrated the potential of directly modifying images by adjusting the noise distribution in diffusion models \citep{xie2023dreaminpainter, meng2022sdedit}. This method is particularly promising as it does not require retraining of the original model.

\textbf{Text guided image editing.}
As the performance of text-to-image diffusion models improves, their use in text-guided image editing has shown significant advantages. This paper aims to address the task of text-guided image editing using the original image and target prompt text as inputs. Currently, most pretrained diffusion models \citep{balaji2022ediff, hoogeboom2023simple, ramesh2022hierarchical, rombach2022high} employ Classifier-Free Guidance (CFG) \citep{ho2022classifier} to enhance conditioned synthesis results. However, due to the semantic space differences between images and text, these models often encounter various semantic misalignment issues where the image does not match the objects and their attributes described in the provided text prompts \citep{conwell2022testing, rassin2022dalle, saharia2022photorealistic, yu2022scaling,diffusioncondition_fair}. 

Prompt-to-Prompt \citep{hertz2022prompt}, by reusing the cross-attention maps of the original image to control text annotations in the edited image, can maintain scene consistency. However, when making significant textual modifications (e.g., changing "bicycle" to "car"), it may lead to geometric inconsistencies between the new and original elements. Text inversion \citep{gal2022image} guides personalized creation by learning new vocabularies in the embedding space with minimal images. DreamBooth \citep{ruiz2023dreambooth} advances this by fine-tuning the model to generate content driven by specific themes, enhancing the personalization features of the images. Imagic \citep{kawar2023imagic} combines the strategies of text inversion and DreamBooth, implementing complex image editing through a three-stage optimization process. However, its approach is slow in execution and prone to overfitting. This paper introduces a streamlined editing method where target features are identified using text alignment and erased with Classifier-Free Guidance (CFG), enhancing efficiency and reducing semantic misalignment between images and text.

\textbf{Concept unlearning in diffusion models.}
In large-scale image generation diffusion models, specific concepts can be removed by fine-tuning the weights of the U-Net \citep{ronneberger2015u} or editing its cross-attention mechanisms \citep{gandikota2023erasing, kumari2023ablating, gandikota2024unified, zhang2023forget}. However, updating the parameters of the U-Net may lead to a decrease in generation quality under unconditional settings. Recent work, inspired by text inversion \citep{gal2022image}, modifies the parameters of the text encoder using a few images of the target concept to remove that concept \citep{fuchi2024erasing}. This paper extends the concept of unlearning to text-guided image editing, employing Classifier-Free Guidance to precisely modify specific image features based on textual descriptions.

\section{Preliminary}
\label{main-pre}

The Latent Diffusion Model (LDM) can be interpreted as a sequence of denoising models for image generation \citep{ldm}. These denoising models learn to predict the noise $\epsilon_\theta  (z_t, c, t) $ is added to the latent variable $z_t$, conditioned on both the current timestep $t$ and the text condition $c$. It is common to use Classifier-Free Guidance (CFG) \citep{diffusion-c-free} necessitates modelling both the conditional and unconditional scores of the diffusion model:
\begin{equation}
\begin{aligned}
\label{eq2}
\overline{\boldsymbol{\epsilon}}_\theta^{(t)}\left(\mathbf{z}_t\right) =\boldsymbol{\epsilon}_\theta^{(t)}\left(\mathbf{z}_t, c\right)+w(\boldsymbol{\epsilon}_\theta^{(t)}\left(\mathbf{z}_t, c\right)-\boldsymbol{\epsilon}_\theta^{(t)}\left(\mathbf{z}_t\right)),
\end{aligned}
\end{equation}
$w$ is a scale for conditioning. In image-to-image tasks, the inference process begins with a Gaussian noise $\mathbf{z}_T$ that is derived from the input image \citep{ldm}. This initial noise is then progressively denoised using the predicted noise component $\overline{\boldsymbol{\epsilon}}_\theta^{(t)}\left(\mathbf{z}_t, c\right)$, which is conditioned on the current latent variable $\mathbf{z}_t$ and the input conditions $c$. Based on Tweedie’s formula \citep{efron2011tweedie} and the reparametrization trick of \citep{ddim}, some concept unlearning works introduce a time-varying noising process if they intend to remove concept $c$ from latent variable $z_t$ with negative guided noise \citep{gandikota2023erasing}:
\begin{equation}
\begin{aligned}
\label{eq3}
\overline{\boldsymbol{\epsilon}}_\theta^{(t)}\left(\mathbf{z}_t\right) =\boldsymbol{\epsilon}_\theta^{(t)}\left(\mathbf{z}_t, c\right)-\eta(\boldsymbol{\epsilon}_\theta^{(t)}\left(\mathbf{z}_t, c\right)-\boldsymbol{\epsilon}_\theta^{(t)}\left(\mathbf{z}_t\right)),
\end{aligned}
\end{equation}
$\eta$ is a scale for unlearning.

\section{Method}
\label{main-method}
In our paper, we have addressed the challenge of modality gaps about accurately locating the specific concept in an image that needs to be modified when performing text-guided instruction in the diffusion method. To address the challenges posed by the inherent complexity of image scenes, we propose a learning and forgetting strategy (LaF) based on the Stable Diffusion model. The key idea is to first identify the specific conceptual elements within the image that require editing, and then selectively forget the original factors associated with those elements. Specifically, we design a localization tool that effectively identifies the semantic information associated with the target location mentioned in the input prompt, and then we employ the forgetting process to give negative guidance in the extracted semantic information.

\begin{figure}[h]
    \centering
    \includegraphics[width=0.98\textwidth]{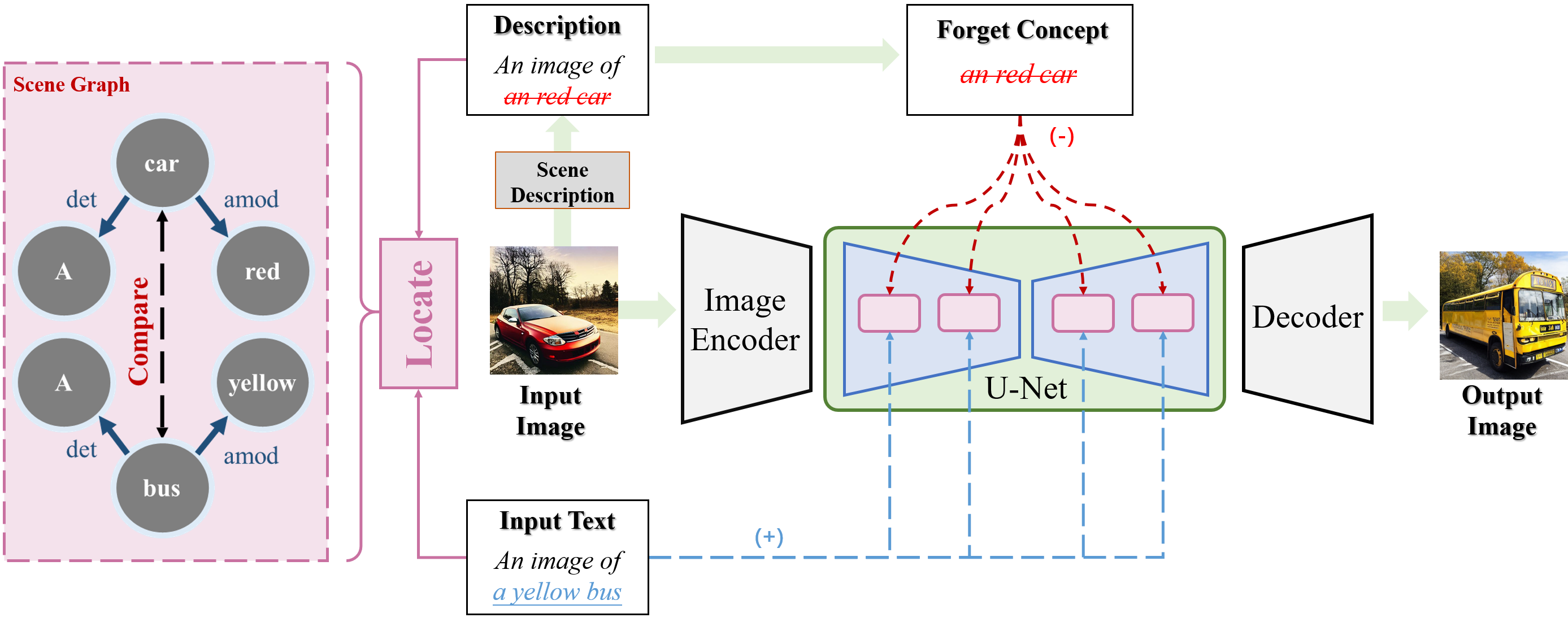}
    \caption{Our framework for our method Locate and Forget (LaF). LaF consists of two parts: 1) Alignment of the input text prompt with the visual scene information: By comparing the textual instructions to the scene description of the image's contents, the LaF model can identify the specific concepts and attributes in the visual scene that need to be edited. 2) Selective forgetting during the diffusion process: During the denoising steps, identified forgettable elements as a form of negative guidance to be removed, which allows to selectively forget the influence of the visual elements that are not aligned with the user's intent.}
    \label{fig:fw}
\end{figure}
\vspace{-10pt}

\subsection{Locating Concepts In Image Context}


The first step in LaF is to accurately locate the regions or components of the image that need to be modified. Image scenes are very complex, containing numerous objects and detailed descriptions. This complexity poses a significant challenge in image editing tasks, as the model needs to have a comprehensive understanding of the various elements within the scene in order to make appropriate and coherent edits. To successfully edit an image, the model must have the capability to recognize and reason about the diverse elements present, such as the spatial relationships between objects, their attributes and functionalities \citep{song2023bridge}, as well as the overall scene context. Therefore, we propose a method to define where edits should be applied within the image. By first establishing a clear understanding of the various semantic elements and their relationships in the scene,  can more effectively identify the appropriate regions to target for editing. 

Firstly, we generate textual captions about the images to give a description. To accomplish this, we applied an image-to-text tool OpenFlamingo \citep{awadalla2023openflamingo}, which is capable of generating comprehensive text descriptions that capture the various concepts with their relationship and status in the image. The scene description serves as a crucial intermediate representation, bridging the gap between the raw visual input and the high-level understanding required for targeted image editing. 

The natural language structure of the captions allows us to obtain syntactic parse trees for each sentence. By analyzing the parse tree, we can identify the thematic elements at the core of each sentence. This syntactic information provides valuable insights into the image's semantic content, which can then be used to guide the subsequent text-to-image editing process. As Figure \ref{fig:fw} shows, the textual description generated from the image, such as "a red car" serves as a main body in the image to provide a detailed description of what the image depicts. In contrast, the input prompt "a yellow bus" expresses the user's intention to replace the red car with a yellow bus. 

We employ dependency parsing to identify the subject of each sentence by spaCy. We consider the root node of the syntactic chunk to be the primary entities depicted in the image, while the children of the root node represent modifiers that describe the specific attributes or state of that object. As the Algorithm \ref{alg:node} shows, by comparing the subject entities mentioned in the image caption and the user's prompt, we can very clearly observe the changes in the primary subjects and themes depicted in the images. If the main subjects differ, we analyze whether the new object introduced in the prompt conflicts with the one described in the image. Conversely, if the subjects are the same, we examine whether the properties in the same entity, such as color (eg. red) and status (eg. standing) have changed between the scene description of the input image and the target prompt.

\begin{algorithm}[H]
    \caption{Locating Algorithm based on Dependency Parsing}
    \label{alg:node}
    \begin{algorithmic}[1]
        \Require Original image $\mathbf{I}$, input prompt $\mathbf{P}$
        \Ensure Figure out the user's editing intent
        \State Perform dependency parsing on $\mathbf{I}$ and $\mathbf{P}$ to from the subject entity chunk sets $C_{i}$ and $C_{p}$ respectively
        \State Initialize $\text{forgetting\_elements} = []$
        \State $\text{common\_entity} = \text{get\_common\_chunk}(C_{i}, C_{p})$
        \If{\text{common\_entity} is empty}
            \State $\text{forgetting\_elements}=C_{i}$
        
        \Else 
            \For{e in common\_entity}
                \State i,p = get\_chunk(e,$C_{i}$),get\_chunk(e,$C_{p}$)
                \State \text{forgetting\_elements}+=(p.children-i.children)
            \EndFor
        \EndIf
    \State \textbf{return} forgetting\_elements
    \end{algorithmic}
\end{algorithm}

\vspace{-10pt}

This detailed analysis of the syntactic structure allows us to precisely pinpoint the focal points of the image and the user's editing intent. We can then use this information to guide the image editing process, ensuring that any modifications are directly targeted at the relevant objects and their associated characteristics.

\subsection{Forgetting Concept for Conditionings}

The primary goal of our method is to forget specific concepts from diffusion models using only the model's existing knowledge in the inference stage, without requiring any additional external data. Classifier-free diffusion guidance is commonly employed in class-conditional and text-conditional image generation tasks to enhance the visual quality of the generated images and to ensure that the sampled outputs better correspond to their respective conditioning factors. Typically, diffusion models use a guidance signal, where the model is conditioned on the desired concept or attribute to generate samples aligned with that concept \citep{gandikota2024unified}. 

Our approach employs negative guidance to allow the diffusion model to gradually forget a specified concept in the original image while applying positive guidance for learning text prompts. Combining with Eq \ref{eq2} and Eq \ref{eq3}, we compute the following score estimate using Classifier-Free Guidance during inference:

\begin{equation}
\begin{aligned}
\label{eq4}
\overline{\boldsymbol{\epsilon}}_\theta^{(t)}\left(\mathbf{z}_t\right)=\boldsymbol{\epsilon}_\theta^{(t)}\left(\mathbf{z}_t\right) & +w(\boldsymbol{\epsilon}_\theta^{(t)}\left(\mathbf{z}_t, c_p\right)-\boldsymbol{\epsilon}_\theta^{(t)}\left(\mathbf{z}_t\right))\\
&-\eta(\boldsymbol{\epsilon}_\theta^{(t)}\left(\mathbf{z}_t, c_n\right)-\boldsymbol{\epsilon}_\theta^{(t)}\left(\mathbf{z}_t\right)),
\end{aligned}
\end{equation}

where $c_p$ and $c_n$ are input prompt and forgetting concepts, $w$ and $\eta$ are weights for controlling the balance between the positive and negative guidance signals in the final CFG score estimate. A higher value of eta will give more weight to the negative guidance, allowing tuning the level of concept forgetting during the diffusion process.

\section{Experiments}
\label{main-exp}
This section presents a series of carefully designed experiments with comprehensive results and detailed ablation studies. Compared to state-of-the-art image editing models HIVE \citep{zhang2023hive}, InstructPix2Pix(IP2P) \citep{IP2P}, and Stable Diffusion Image-to-Image(SD) \citep{ldm}, our work makes the following contributions: 1) We comprehensively assess our model's alignment and quality capabilities. 2) We perform detailed ablation studies to analyze the impact of forgetting scales within our model architecture. 3) We also conduct a human preference investigation, soliciting user feedback on the output images.

\subsection{Experimental Setup}

\begin{figure*}[t]
\centering
\includegraphics[width=0.9\textwidth]{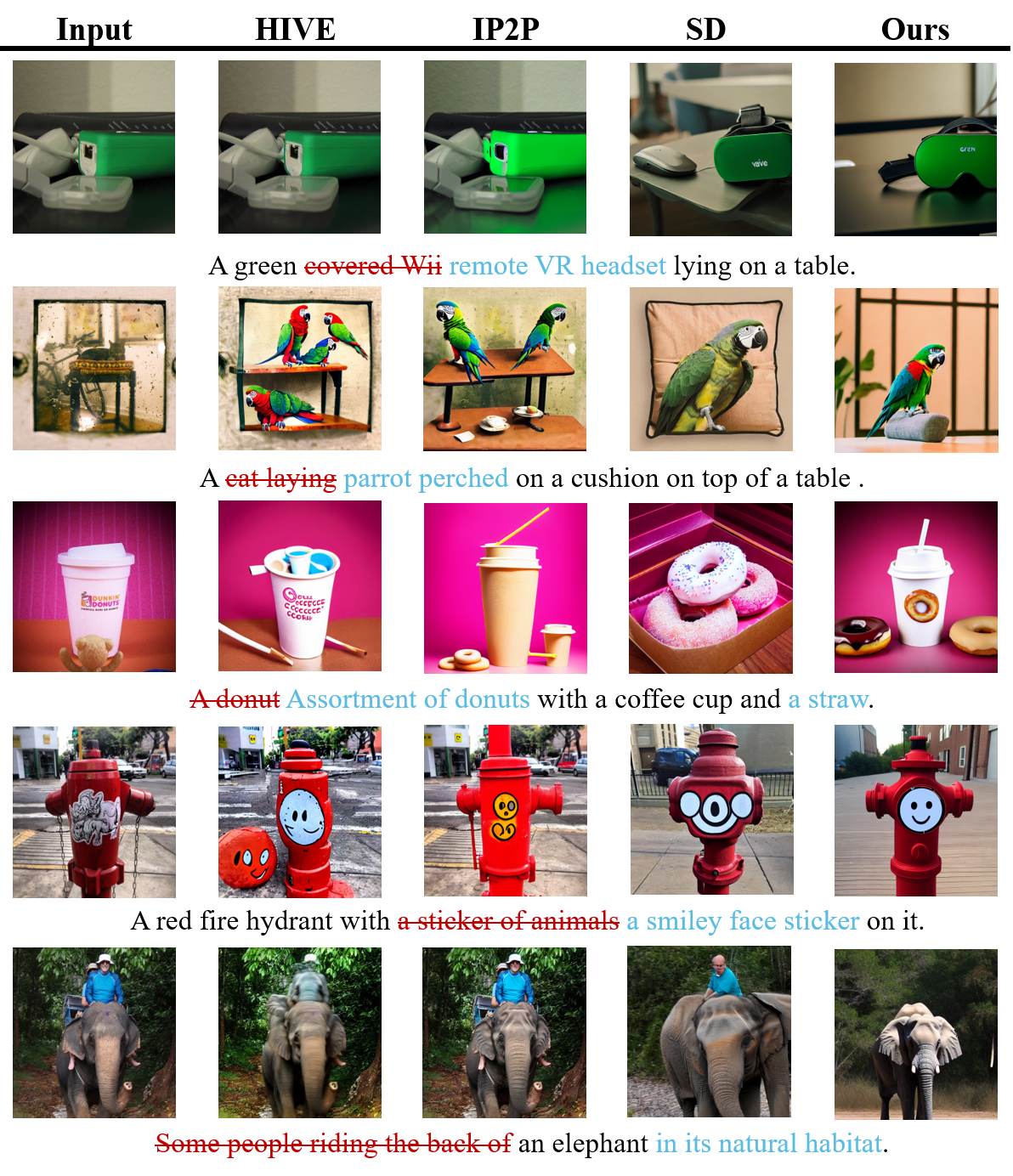}
\vspace{-10pt}
\caption{Visual comparisons between Hive, IP2P, SD and our method in dataset MagicBrush. The red annotations indicate the visual concepts in the original image that need to be edited, while the blue annotations represent the new visual elements that should be introduced based on the provided textual prompt.}
\label{fig:exp-1}
\vspace{-10pt}
\end{figure*}

\textbf{Datasets.} To evaluate our model's capabilities in image editing, we consider two benchmark tasks: TedBench and MagicBrush. 1) TedBench\citep{kawar2023imagic}: a public-available text-guided image editing benchmark, consists of 100 editing tasks, each focusing on a single, clearly defined concept within the image. 2) MagicBrush\citep{zhang2024magicbrush}: a synthetic image-editing benchmark, containing a diverse range of complex real scenarios. we specifically select 545 samples with a single text prompt for editing. 3) Human Prompt Generated Dataset: a human-written dataset of editing triplets from \citet{dataset-laion5b}, where we used 700 textual prompts about Aesthetics to generate 700 corresponding images by the Stable Diffusion text-to-image model and then perform editing operations on each image. Details for datasets are provided in the supplementary materials.

\textbf{Implementation details.} In our experiments, we use Stable Diffusion \citep{ldm} as the base model and implement 50 DDIM denoising steps for generation. Specifically, we utilize the pre-trained stable diffusion model (SD-1.5). All of our training experiments are conducted on one Nvidia V100 GPU. For the key hyperparameters, we set the $w$ (weight) to 10 and the $\eta$ (guidance scale) to 2.5. This configuration was chosen to strike a balance with the guidance scale setting of 7.5 commonly used in Stable Diffusion as well as the other models we compared in this paper.

\textbf{Evaluation metrics.} To evaluate the quality of the generated images, we utilized the Inception Score (IS) \citep{IS} metric, which provides a quantitative assessment of how realistic the generated images appear by measuring the conditional class probabilities predicted by a pre-trained Inception network. CLIP-T \citep{metri_clip_score} is applied to measure the similarity between the textual descriptions and the visual content of the generated images. We take L1 as the metric to evaluate the distance between real images and generated results. 

Despite the high pixel-level overlap between the generated images and the original images, a disconnect in the semantic cohesion appears\citep{zhang2023hive}. To quantify the degree of semantic change induced by our image editing approach, we propose measures for the difference of the semantic distance between the generated images and their edited versions images, as well as the semantic distance between the generated images and the original unedited inputs, which can represent if generative models just replicate the low-level visual features in input images. More details are presented in the Appendix \ref{clip-d}.

\subsection{Main Experiments}

\vspace{-5pt}
\begin{table}[h]
\centering
\caption{Comparison of various metrics across different datasets. The lack of Human Prompt Generated Dataset dataset with before-and-after images hinders the evaluation of text-guided image editing models using metrics like CLIP-D.} \label{tab:exp1}
\resizebox{0.9\linewidth}{!}{
\begin{tabular}{@{}llcccc@{}}
\toprule
\textbf{Dataset} & \textbf{Metric} & \textbf{LaF (Ours)} & \textbf{SD} & \textbf{IP2P} & \textbf{HIVE} \\
\midrule
\multirow{4}{*}{\textbf{TedBench}} & CLIP-T & \textbf{28.94} & 28.35 & 28.18 & 28.11 \\
                          & CLIP-D & \textbf{4.13e-02} & 2.79e-02 & -4.03e-02 & -4.43e-02 \\
                          & IS & \textbf{18.69} & 18.62 & 15.81 & 16.66 \\
                          & L1 & 0.51 & 0.53 & \textbf{0.32} & 0.38 \\
\midrule
\multirow{4}{*}{\textbf{MagicBrush}} & CLIP-T & \textbf{26.72} & 26.23 & 25.87 & 25.04 \\
                            & CLIP-D & \textbf{2.63e-02} & 1.00e-04 & 3.00e-04 & -6.42e-03 \\
                            & IS & 35.21 & \textbf{35.40} & 35.38 & 30.81 \\
                            & L1 & 0.72 & 0.76 & 0.61 & \textbf{0.53} \\
\midrule
\multirow{4}{*}{\textbf{Human Prompt Generated Dataset}} & CLIP-T & \textbf{24.46} & 24.06 & 24.19 & 23.27 \\
                                                & CLIP-D & - & - & - & - \\
                                                & IS & 19.49 & 19.33 & 19.74 & \textbf{19.75} \\
                                                & L1 & 0.56 & 0.60 & 0.50 & \textbf{0.29} \\
\bottomrule
\end{tabular}}
\end{table}

\vspace{-5pt}

\textbf{Visual results.} In figure \ref{fig:exp-1}, our proposed method demonstrates the ability to generate more natural and visually compelling images compared to existing approaches on the dataset MagicBrush. The generated images exhibit enhanced realism and better alignment with the semantic concepts described in the textual conditions. Considering an image depicting "some people riding the back of an elephant." when given the editing prompt "An elephant in its natural habitat.", our text-guided image editing approach can successfully address this task. LaF successfully removes the people from the original image of an elephant, preserves the authentic appearance of the elephant, and generates an edited image that faithfully as specified by the textual prompt, while other methods can only duplicate existing visual content. More visual results are presented in Appendix \ref{app-visual}.


\textbf{Alignment.} We report the quantitative results about alignment between the generated images and their corresponding textual condition in the Three evaluation datasets in Table \ref{tab:exp1}. Our method has demonstrated promising results, both in synthetic image datasets and real-world image datasets compared to other existing methods. Specifically, our model achieved a CLIP-T value up to 1.68 points higher than the results generated by other methods. This notable improvement in performance highlights the effectiveness of the superior ability of our text-guided editing method to generate edited images that closely align with the provided textual instructions.

\textbf{Quality of generated images.} Based on the provided data in Table \ref{tab:exp1}, we conducted a brief analysis of the Inception Score (IS) metric. Overall, our proposed LaF method demonstrated the strongest performance, achieving the highest IS score of 18.69 on the real-world dataset TedBench dataset, outperforming the other approaches. Moreover, LaF also demonstrated robust and consistent performance among synthetic datasets.

\textbf{Consistency.} 
As for the L1 loss calculated between the edited and original images, the HIVE model demonstrated a significant advantage, exhibiting up to a 48\% lower loss compared to other methods, which indicates that the HIVE approach can preserve a larger portion of the original image details during the editing process. While the other models have all shown the capability to preserve a substantial amount of the original content. However, the significantly lower L1 loss also raises a potential concern that the preservation of original image details also indicates the copying or reproducing of large portions of the input image, rather than actively applying meaningful edits based on the provided textual guidance, as shown in Figure \ref{fig:exp-1}.


We measure the difference in editing scales by CLIP-D, which evaluates the direction of the image editing action in the intended semantic content. A larger positive CLIP-D score demonstrates that the image editing techniques are more successfully translating the semantic intent of the text into corresponding visual editing. Conversely, a negative CLIP-D score implies that the generation reserves the input image rather than the target semantics defined in textual prompts. It can be seen that on the real image dataset TedBench, the CLIP-D results for HIVE and IP2P are even under 0, which indicates that the contents of the generated images create duplicate input images, rather than align with the provided text instructions. Instead, our model leads to edits more consistent with the target text descriptions, which shows a positive result in semantic content editing. We did not conduct CLIP-D evaluations on the Human Prompt Generated Dataset, as this dataset consists of images generated from human-written prompts, without corresponding modified versions of the images.

\subsection{Ablation Study}

\begin{figure*}[ht]
\centering
\includegraphics[width=0.95\textwidth]{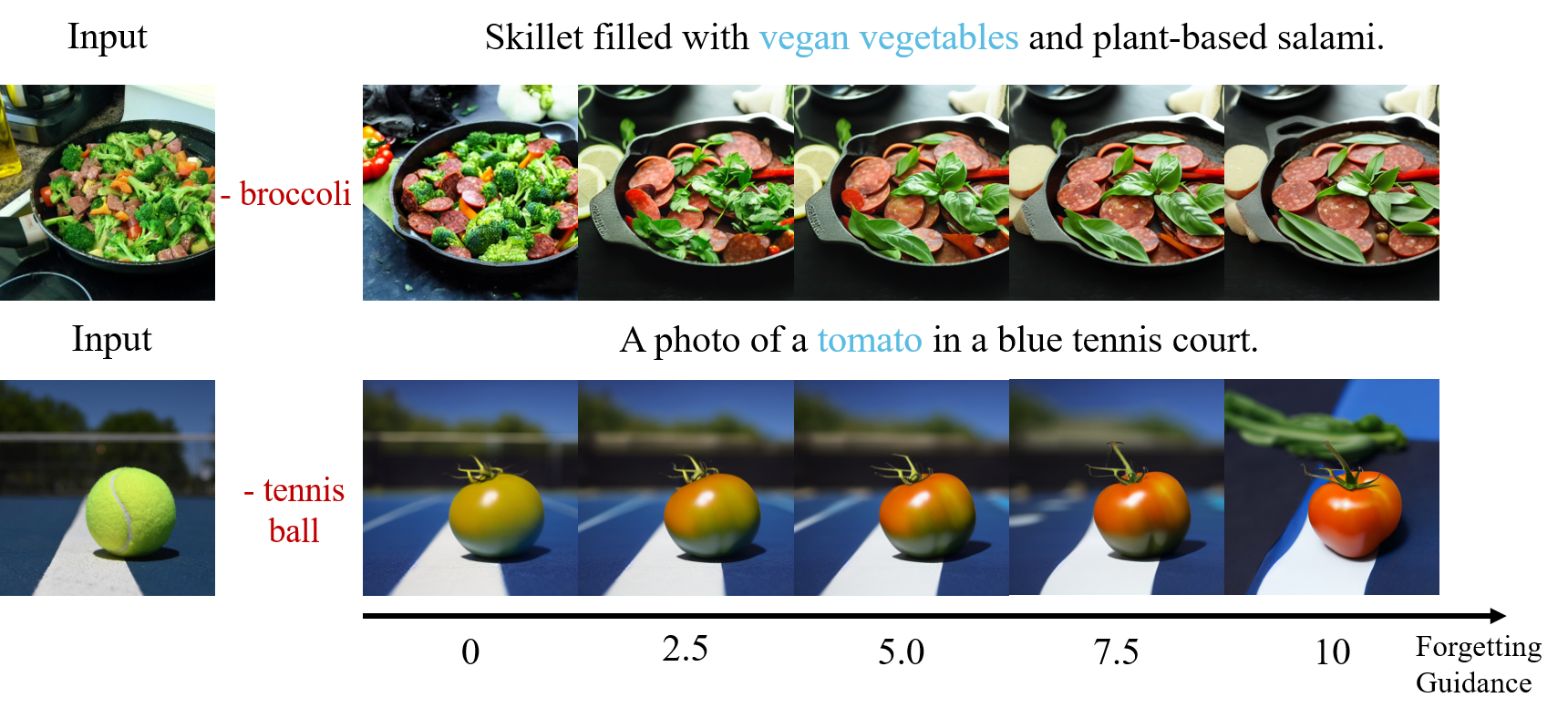}
\vspace{-5pt}
\caption{Visual editing results under Varying Forgetting Guidance $\eta$. Example images are respectively \textit{"Skillet filled with salami, broccoli and other vegetables."} and \textit{"A tennis ball on a tennis court."}}
\label{fig:exp-2}
\end{figure*}
\vspace{-5pt}

The visualization of our experiments in Figure \ref{fig:exp-2} varying the forgetting guidance coefficient provides clear insights into the impact of forgetting guidance $\eta$ on the generated images, where we fix $w$ as $7.5$. As the forgetting guidance coefficient is increased, the generating images exhibit a more pronounced degree of forgetting. For instance, when attempting to edit the tennis ball in the original figure into a tomato, the increased forgetting guidance leads to the generated tomato retaining some of the original green appearances of the tennis ball. However, as the forgetting coefficient is further amplified, the resulting tomatoes become increasingly closer to their typical red color. It indicates that our text-guided image editing methods can successfully remove certain semantic attributes present in the original image and avoid bad influence in editing.

We conducted experiments across three diverse datasets to investigate the impact of different forgetting guidance values on the CLIP-T in Figure \ref{fig:exp-ab1}. Our findings reveal that overall, a mild degree of forgetting guidance (less than 5) can lead to substantial improvements in alignment, with the optimal performance achieved at a forgetting guidance value of 2.5. Meanwhile, as the forgetting guidance is increased beyond a certain threshold, the edited images begin to lose their semantic coherence, resulting in a rapid decline in CLIP scores. Interestingly, our method exhibited the strongest performance on the real-world images dataset TedBench, outperforming the results on the Human Prompt Generated Dataset by an average of 18\% and surpassing MagicBrush by 8\%, which suggests our approach is particularly effective at maintaining high semantic alignment on more natural and diverse visual data. Additionally, the MagicBrush dataset is the most sensitive to forgetting guidance, with semantic degradation occurring at lower levels than other datasets, emphasizing the importance of carefully tuning the forgetting guidance based on target dataset characteristics.

\begin{minipage}{\textwidth}
\begin{minipage}[ht]{0.48\textwidth}
\makeatletter\def\@captype{figure}
\centering
\includegraphics[width=0.95\textwidth]{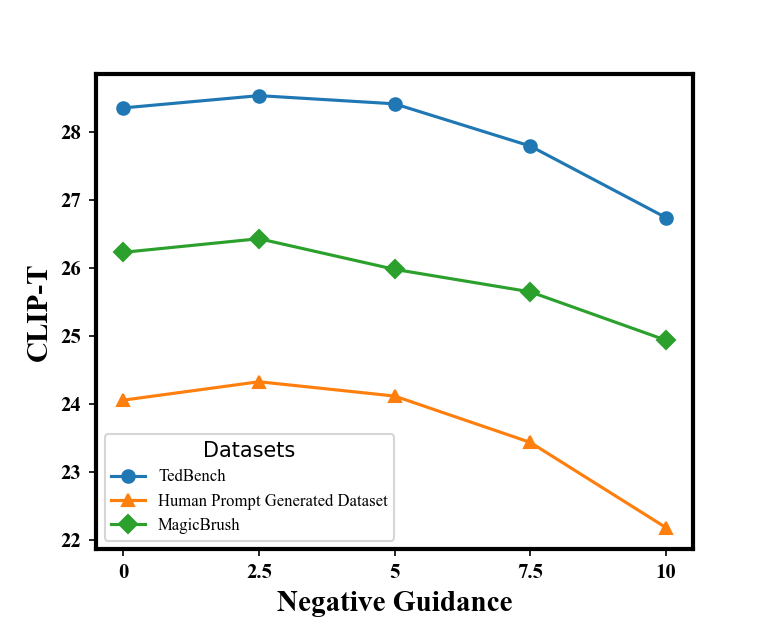}
\caption{Impact of Forgetting Guidance values on CLIP-T across different Datasets}\label{fig:exp-ab1}
\end{minipage}
\begin{minipage}[ht]{0.48\textwidth}
\makeatletter\def\@captype{figure}
\centering
\includegraphics[width=0.95\textwidth]{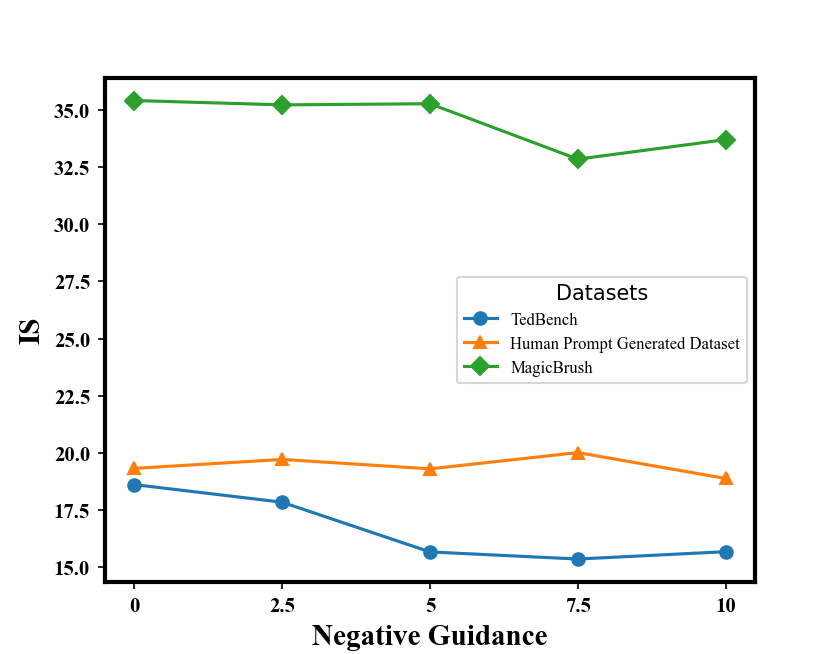}
\caption{Impact of Forgetting Guidance values on IS across different Datasets}\label{fig:exp-ab2}
\end{minipage}
\end{minipage}
\vspace{3pt}

The experimental results in Figure \ref{fig:exp-ab2} show that the Inception Score (IS) values exhibit fluctuations as the Forgetting Guidance $\eta$ increases. However, when the Forgetting Guidance becomes very large, the IS performance drops significantly. It would be advisable to recommend using a comparatively small Forgetting Guidance value to maintain high-quality image generation.

\subsection{Human Preference} 

The provided table in table \ref{tb:exp2} presents a comparative evaluation of different image generation methods across four key metrics: alignment, fidelity, consistency, and overall human preferability. Our LaF method outperforms the other approaches across multiple dimensions. According to the human preference metric, LaF achieves the highest rating of 3.42. In detail, LaF achieves the best performance in terms of both Fidelity and alignment. The results of the user study are consistent with our previous analysis. The survey findings indicate that our proposed method, LaF, is generally considered to have stronger alignment and better editing outcomes compared to the other approaches evaluated. It also avoids the issue of generating repetitive results of the image content.

\begin{table}[h]
\centering
\caption{Alignment Fidelity, Consistency, and Preferability Across Methods. Higher values means better. More details in Appendix.} \label{tb:exp2}
\begin{tabular}{@{}lcccc@{}}
\toprule
\textbf{Method} & \textbf{Alignment} & \textbf{Fidelity} & \textbf{Consistency} & \textbf{Preferable} \\
\midrule
LaF (Ours)   & 3.67 & 3.44 & 3.48 & 3.42 \\
SD & 3.44 & 3.31 & 3.29 & 3.01 \\
IP2P   & 3.49 & 3.35 & 3.51 & 3.39 \\
HIVE   & 3.31 & 3.41 & 3.46 & 3.23 \\
\bottomrule
\end{tabular}
\end{table}

\vspace{-10pt}

\section{Conclusion}
The challenge of seamlessly incorporating objects into images based on textual instructions without relying on extra user-provided guidance remains. In our paper, we introduce a novel method, Locating and Forgetting, to improve text-guided image editing. By constructing a scene description from the image's visual concepts and comparing the syntactic parse tree to the textual prompt, our method can precisely locate the relevant concepts and forget only those concepts based on the provided textual conditions, which bridges the gap between textual instructions and visual reality. Through extensive experimentation, we demonstrate that our method significantly enhances the alignment and quality of editing results rather than simply replicating the input content.

\clearpage
\bibliographystyle{abbrvnat}
\bibliography{reference}

\begin{thebibliography}{52}
\providecommand{\natexlab}[1]{#1}
\providecommand{\url}[1]{\texttt{#1}}
\expandafter\ifx\csname urlstyle\endcsname\relax
  \providecommand{\doi}[1]{doi: #1}\else
  \providecommand{\doi}{doi: \begingroup \urlstyle{rm}\Url}\fi

\bibitem[Avrahami et~al.(2022)Avrahami, Lischinski, and Fried]{avrahami2022blended}
O.~Avrahami, D.~Lischinski, and O.~Fried.
\newblock Blended diffusion for text-driven editing of natural images.
\newblock In \emph{Proceedings of the IEEE/CVF Conference on Computer Vision and Pattern Recognition}, pages 18208--18218, 2022.

\bibitem[Awadalla et~al.(2023)Awadalla, Gao, Gardner, Hessel, Hanafy, Zhu, Marathe, Bitton, Gadre, Sagawa, et~al.]{awadalla2023openflamingo}
A.~Awadalla, I.~Gao, J.~Gardner, J.~Hessel, Y.~Hanafy, W.~Zhu, K.~Marathe, Y.~Bitton, S.~Gadre, S.~Sagawa, et~al.
\newblock Openflamingo: An open-source framework for training large autoregressive vision-language models.
\newblock \emph{arXiv preprint arXiv:2308.01390}, 2023.

\bibitem[Balaji et~al.(2022)Balaji, Nah, Huang, Vahdat, Song, Zhang, Kreis, Aittala, Aila, Laine, et~al.]{balaji2022ediff}
Y.~Balaji, S.~Nah, X.~Huang, A.~Vahdat, J.~Song, Q.~Zhang, K.~Kreis, M.~Aittala, T.~Aila, S.~Laine, et~al.
\newblock ediff-i: Text-to-image diffusion models with an ensemble of expert denoisers.
\newblock \emph{arXiv preprint arXiv:2211.01324}, 2022.

\bibitem[Barratt and Sharma(2018)]{IS}
S.~Barratt and R.~Sharma.
\newblock A note on the inception score.
\newblock \emph{arXiv preprint arXiv:1801.01973}, 2018.

\bibitem[Brooks et~al.(2023)Brooks, Holynski, and Efros]{IP2P}
T.~Brooks, A.~Holynski, and A.~A. Efros.
\newblock Instructpix2pix: Learning to follow image editing instructions, 2023.

\bibitem[Conwell and Ullman(2022)]{conwell2022testing}
C.~Conwell and T.~Ullman.
\newblock Testing relational understanding in text-guided image generation.
\newblock \emph{arXiv preprint arXiv:2208.00005}, 2022.

\bibitem[Efron(2011)]{efron2011tweedie}
B.~Efron.
\newblock Tweedie’s formula and selection bias.
\newblock \emph{Journal of the American Statistical Association}, 106\penalty0 (496):\penalty0 1602--1614, 2011.

\bibitem[Fuchi and Takagi(2024)]{fuchi2024erasing}
M.~Fuchi and T.~Takagi.
\newblock Erasing concepts from text-to-image diffusion models with few-shot unlearning.
\newblock \emph{arXiv preprint arXiv:2405.07288}, 2024.

\bibitem[Gal et~al.(2022)Gal, Alaluf, Atzmon, Patashnik, Bermano, Chechik, and Cohen-Or]{gal2022image}
R.~Gal, Y.~Alaluf, Y.~Atzmon, O.~Patashnik, A.~H. Bermano, G.~Chechik, and D.~Cohen-Or.
\newblock An image is worth one word: Personalizing text-to-image generation using textual inversion, 2022.

\bibitem[Gandikota et~al.(2023)Gandikota, Materzynska, Fiotto-Kaufman, and Bau]{gandikota2023erasing}
R.~Gandikota, J.~Materzynska, J.~Fiotto-Kaufman, and D.~Bau.
\newblock Erasing concepts from diffusion models.
\newblock In \emph{Proceedings of the IEEE/CVF International Conference on Computer Vision}, pages 2426--2436, 2023.

\bibitem[Gandikota et~al.(2024)Gandikota, Orgad, Belinkov, Materzy{\'n}ska, and Bau]{gandikota2024unified}
R.~Gandikota, H.~Orgad, Y.~Belinkov, J.~Materzy{\'n}ska, and D.~Bau.
\newblock Unified concept editing in diffusion models.
\newblock In \emph{Proceedings of the IEEE/CVF Winter Conference on Applications of Computer Vision}, pages 5111--5120, 2024.

\bibitem[Hertz et~al.(2022)Hertz, Mokady, Tenenbaum, Aberman, Pritch, and Cohen-Or]{hertz2022prompt}
A.~Hertz, R.~Mokady, J.~Tenenbaum, K.~Aberman, Y.~Pritch, and D.~Cohen-Or.
\newblock Prompt-to-prompt image editing with cross attention control.
\newblock \emph{arXiv preprint arXiv:2208.01626}, 2022.

\bibitem[Hessel et~al.(2021)Hessel, Holtzman, Forbes, Bras, and Choi]{metri_clip_score}
J.~Hessel, A.~Holtzman, M.~Forbes, R.~L. Bras, and Y.~Choi.
\newblock Clipscore: {A} reference-free evaluation metric for image captioning.
\newblock In M.~Moens, X.~Huang, L.~Specia, and S.~W. Yih, editors, \emph{Proceedings of the 2021 Conference on Empirical Methods in Natural Language Processing, {EMNLP} 2021, Virtual Event / Punta Cana, Dominican Republic, 7-11 November, 2021}, pages 7514--7528. Association for Computational Linguistics, 2021.
\newblock \doi{10.18653/v1/2021.emnlp-main.595}.
\newblock URL \url{https://doi.org/10.18653/v1/2021.emnlp-main.595}.

\bibitem[Ho and Salimans(2022{\natexlab{a}})]{diffusion-c-free}
J.~Ho and T.~Salimans.
\newblock Classifier-free diffusion guidance.
\newblock \emph{CoRR}, abs/2207.12598, 2022{\natexlab{a}}.
\newblock \doi{10.48550/arXiv.2207.12598}.
\newblock URL \url{https://doi.org/10.48550/arXiv.2207.12598}.

\bibitem[Ho and Salimans(2022{\natexlab{b}})]{ho2022classifier}
J.~Ho and T.~Salimans.
\newblock Classifier-free diffusion guidance.
\newblock \emph{arXiv preprint arXiv:2207.12598}, 2022{\natexlab{b}}.

\bibitem[Ho et~al.(2020{\natexlab{a}})Ho, Jain, and Abbeel]{ddpm}
J.~Ho, A.~Jain, and P.~Abbeel.
\newblock Denoising diffusion probabilistic models.
\newblock In H.~Larochelle, M.~Ranzato, R.~Hadsell, M.~Balcan, and H.~Lin, editors, \emph{Advances in Neural Information Processing Systems 33: Annual Conference on Neural Information Processing Systems 2020, NeurIPS 2020, December 6-12, 2020, virtual}, 2020{\natexlab{a}}.
\newblock URL \url{https://proceedings.neurips.cc/paper/2020/hash/4c5bcfec8584af0d967f1ab10179ca4b-Abstract.html}.

\bibitem[Ho et~al.(2020{\natexlab{b}})Ho, Jain, and Abbeel]{ho2020denoising}
J.~Ho, A.~Jain, and P.~Abbeel.
\newblock Denoising diffusion probabilistic models.
\newblock \emph{Advances in neural information processing systems}, 33:\penalty0 6840--6851, 2020{\natexlab{b}}.

\bibitem[Hoogeboom et~al.(2023)Hoogeboom, Heek, and Salimans]{hoogeboom2023simple}
E.~Hoogeboom, J.~Heek, and T.~Salimans.
\newblock simple diffusion: End-to-end diffusion for high resolution images.
\newblock In \emph{International Conference on Machine Learning}, pages 13213--13232. PMLR, 2023.

\bibitem[Hu et~al.(2023)Hu, Liu, Liu, Huai, Sun, and Wang]{vit}
L.~Hu, Y.~Liu, N.~Liu, M.~Huai, L.~Sun, and D.~Wang.
\newblock Improving faithfulness for vision transformers.
\newblock \emph{arXiv preprint arXiv:2311.17983}, 2023.

\bibitem[Hu et~al.(2024)Hu, Zhang, Yi, Du, Chen, Liu, Wang, and Wang]{hu2024anomalydiffusion}
T.~Hu, J.~Zhang, R.~Yi, Y.~Du, X.~Chen, L.~Liu, Y.~Wang, and C.~Wang.
\newblock Anomalydiffusion: Few-shot anomaly image generation with diffusion model.
\newblock In \emph{Proceedings of the AAAI Conference on Artificial Intelligence}, volume~38, pages 8526--8534, 2024.

\bibitem[Huang et~al.(2024)Huang, Huang, Liu, Yan, Lv, Liu, Xiong, Zhang, Chen, and Cao]{huang2024diffusion}
Y.~Huang, J.~Huang, Y.~Liu, M.~Yan, J.~Lv, J.~Liu, W.~Xiong, H.~Zhang, S.~Chen, and L.~Cao.
\newblock Diffusion model-based image editing: A survey, 2024.

\bibitem[Kawar et~al.(2023)Kawar, Zada, Lang, Tov, Chang, Dekel, Mosseri, and Irani]{kawar2023imagic}
B.~Kawar, S.~Zada, O.~Lang, O.~Tov, H.~Chang, T.~Dekel, I.~Mosseri, and M.~Irani.
\newblock Imagic: Text-based real image editing with diffusion models.
\newblock In \emph{Conference on Computer Vision and Pattern Recognition 2023}, 2023.

\bibitem[Kumari et~al.(2023)Kumari, Zhang, Wang, Shechtman, Zhang, and Zhu]{kumari2023ablating}
N.~Kumari, B.~Zhang, S.-Y. Wang, E.~Shechtman, R.~Zhang, and J.-Y. Zhu.
\newblock Ablating concepts in text-to-image diffusion models.
\newblock In \emph{Proceedings of the IEEE/CVF International Conference on Computer Vision}, pages 22691--22702, 2023.

\bibitem[Lai et~al.(2023)Lai, Hu, Wang, Berti-Equille, and Wang]{vlm}
S.~Lai, L.~Hu, J.~Wang, L.~Berti-Equille, and D.~Wang.
\newblock Faithful vision-language interpretation via concept bottleneck models.
\newblock In \emph{The Twelfth International Conference on Learning Representations}, 2023.

\bibitem[Li et~al.(2022)Li, Li, Xiong, and Hoi]{blip}
J.~Li, D.~Li, C.~Xiong, and S.~C.~H. Hoi.
\newblock {BLIP:} bootstrapping language-image pre-training for unified vision-language understanding and generation.
\newblock \emph{CoRR}, abs/2201.12086, 2022.
\newblock URL \url{https://arxiv.org/abs/2201.12086}.

\bibitem[Li et~al.(2023{\natexlab{a}})Li, Hu, Zhang, Zheng, Zhang, and Wang]{diffusioncondition_fair}
J.~Li, L.~Hu, J.~Zhang, T.~Zheng, H.~Zhang, and D.~Wang.
\newblock Fair text-to-image diffusion via fair mapping.
\newblock \emph{arXiv preprint arXiv:2311.17695}, 2023{\natexlab{a}}.

\bibitem[Li et~al.(2023{\natexlab{b}})Li, Liu, Wu, Mu, Yang, Gao, Li, and Lee]{li2023gligen}
Y.~Li, H.~Liu, Q.~Wu, F.~Mu, J.~Yang, J.~Gao, C.~Li, and Y.~J. Lee.
\newblock Gligen: Open-set grounded text-to-image generation.
\newblock In \emph{Proceedings of the IEEE/CVF Conference on Computer Vision and Pattern Recognition}, pages 22511--22521, 2023{\natexlab{b}}.

\bibitem[Liu et~al.(2024)Liu, Wang, Cao, Jia, and Huang]{liu2024understanding}
B.~Liu, C.~Wang, T.~Cao, K.~Jia, and J.~Huang.
\newblock Towards understanding cross and self-attention in stable diffusion for text-guided image editing, 2024.

\bibitem[Liu et~al.(2023)Liu, Li, Wu, and Lee]{liu2023visual}
H.~Liu, C.~Li, Q.~Wu, and Y.~J. Lee.
\newblock Visual instruction tuning, 2023.

\bibitem[Meng et~al.(2022)Meng, He, Song, Song, Wu, Zhu, and Ermon]{meng2022sdedit}
C.~Meng, Y.~He, Y.~Song, J.~Song, J.~Wu, J.-Y. Zhu, and S.~Ermon.
\newblock Sdedit: Guided image synthesis and editing with stochastic differential equations, 2022.

\bibitem[Nichol et~al.(2021)Nichol, Dhariwal, Ramesh, Shyam, Mishkin, McGrew, Sutskever, and Chen]{nichol2021glide}
A.~Nichol, P.~Dhariwal, A.~Ramesh, P.~Shyam, P.~Mishkin, B.~McGrew, I.~Sutskever, and M.~Chen.
\newblock Glide: Towards photorealistic image generation and editing with text-guided diffusion models.
\newblock \emph{arXiv preprint arXiv:2112.10741}, 2021.

\bibitem[Park et~al.(2024)Park, Luo, Toste, Azadi, Liu, Karalashvili, Rohrbach, and Darrell]{Park_2024_WACV}
D.~H. Park, G.~Luo, C.~Toste, S.~Azadi, X.~Liu, M.~Karalashvili, A.~Rohrbach, and T.~Darrell.
\newblock Shape-guided diffusion with inside-outside attention.
\newblock In \emph{Proceedings of the IEEE/CVF Winter Conference on Applications of Computer Vision (WACV)}, pages 4198--4207, January 2024.

\bibitem[Patashnik et~al.(2023)Patashnik, Garibi, Azuri, Averbuch-Elor, and Cohen-Or]{patashnik2023localizing}
O.~Patashnik, D.~Garibi, I.~Azuri, H.~Averbuch-Elor, and D.~Cohen-Or.
\newblock Localizing object-level shape variations with text-to-image diffusion models, 2023.

\bibitem[Podell et~al.(2023)Podell, English, Lacey, Blattmann, Dockhorn, M{\"u}ller, Penna, and Rombach]{podell2023sdxl}
D.~Podell, Z.~English, K.~Lacey, A.~Blattmann, T.~Dockhorn, J.~M{\"u}ller, J.~Penna, and R.~Rombach.
\newblock Sdxl: Improving latent diffusion models for high-resolution image synthesis.
\newblock \emph{arXiv preprint arXiv:2307.01952}, 2023.

\bibitem[Ramesh et~al.(2022)Ramesh, Dhariwal, Nichol, Chu, and Chen]{ramesh2022hierarchical}
A.~Ramesh, P.~Dhariwal, A.~Nichol, C.~Chu, and M.~Chen.
\newblock Hierarchical text-conditional image generation with clip latents.
\newblock \emph{arXiv preprint arXiv:2204.06125}, 1\penalty0 (2):\penalty0 3, 2022.

\bibitem[Rassin et~al.(2022)Rassin, Ravfogel, and Goldberg]{rassin2022dalle}
R.~Rassin, S.~Ravfogel, and Y.~Goldberg.
\newblock Dalle-2 is seeing double: flaws in word-to-concept mapping in text2image models.
\newblock \emph{arXiv preprint arXiv:2210.10606}, 2022.

\bibitem[Rombach et~al.(2022{\natexlab{a}})Rombach, Blattmann, Lorenz, Esser, and Ommer]{ldm}
R.~Rombach, A.~Blattmann, D.~Lorenz, P.~Esser, and B.~Ommer.
\newblock High-resolution image synthesis with latent diffusion models.
\newblock In \emph{{IEEE/CVF} Conference on Computer Vision and Pattern Recognition, {CVPR} 2022, New Orleans, LA, USA, June 18-24, 2022}, pages 10674--10685. {IEEE}, 2022{\natexlab{a}}.
\newblock \doi{10.1109/CVPR52688.2022.01042}.
\newblock URL \url{https://doi.org/10.1109/CVPR52688.2022.01042}.

\bibitem[Rombach et~al.(2022{\natexlab{b}})Rombach, Blattmann, Lorenz, Esser, and Ommer]{rombach2022high}
R.~Rombach, A.~Blattmann, D.~Lorenz, P.~Esser, and B.~Ommer.
\newblock High-resolution image synthesis with latent diffusion models.
\newblock In \emph{Proceedings of the IEEE/CVF conference on computer vision and pattern recognition}, pages 10684--10695, 2022{\natexlab{b}}.

\bibitem[Ronneberger et~al.(2015)Ronneberger, Fischer, and Brox]{ronneberger2015u}
O.~Ronneberger, P.~Fischer, and T.~Brox.
\newblock U-net: Convolutional networks for biomedical image segmentation.
\newblock In \emph{Medical image computing and computer-assisted intervention--MICCAI 2015: 18th international conference, Munich, Germany, October 5-9, 2015, proceedings, part III 18}, pages 234--241. Springer, 2015.

\bibitem[Ruiz et~al.(2023)Ruiz, Li, Jampani, Pritch, Rubinstein, and Aberman]{ruiz2023dreambooth}
N.~Ruiz, Y.~Li, V.~Jampani, Y.~Pritch, M.~Rubinstein, and K.~Aberman.
\newblock Dreambooth: Fine tuning text-to-image diffusion models for subject-driven generation, 2023.

\bibitem[Saharia et~al.(2022)Saharia, Chan, Saxena, Li, Whang, Denton, Ghasemipour, Gontijo~Lopes, Karagol~Ayan, Salimans, et~al.]{saharia2022photorealistic}
C.~Saharia, W.~Chan, S.~Saxena, L.~Li, J.~Whang, E.~L. Denton, K.~Ghasemipour, R.~Gontijo~Lopes, B.~Karagol~Ayan, T.~Salimans, et~al.
\newblock Photorealistic text-to-image diffusion models with deep language understanding.
\newblock \emph{Advances in neural information processing systems}, 35:\penalty0 36479--36494, 2022.

\bibitem[Schuhmann et~al.(2022)Schuhmann, Beaumont, Vencu, Gordon, Wightman, Cherti, Coombes, Katta, Mullis, Wortsman, Schramowski, Kundurthy, Crowson, Schmidt, Kaczmarczyk, and Jitsev]{dataset-laion5b}
C.~Schuhmann, R.~Beaumont, R.~Vencu, C.~Gordon, R.~Wightman, M.~Cherti, T.~Coombes, A.~Katta, C.~Mullis, M.~Wortsman, P.~Schramowski, S.~Kundurthy, K.~Crowson, L.~Schmidt, R.~Kaczmarczyk, and J.~Jitsev.
\newblock {LAION-5B:} an open large-scale dataset for training next generation image-text models.
\newblock In \emph{NeurIPS}, 2022.
\newblock URL \url{http://papers.nips.cc/paper\_files/paper/2022/hash/a1859debfb3b59d094f3504d5ebb6c25-Abstract-Datasets\_and\_Benchmarks.html}.

\bibitem[Song et~al.(2021)Song, Meng, and Ermon]{ddim}
J.~Song, C.~Meng, and S.~Ermon.
\newblock Denoising diffusion implicit models.
\newblock In \emph{9th International Conference on Learning Representations, {ICLR} 2021, Virtual Event, Austria, May 3-7, 2021}. OpenReview.net, 2021.
\newblock URL \url{https://openreview.net/forum?id=St1giarCHLP}.

\bibitem[Song et~al.(2023)Song, Li, Li, Zhao, Yu, Ma, Mao, and Zhang]{song2023bridge}
S.~Song, X.~Li, S.~Li, S.~Zhao, J.~Yu, J.~Ma, X.~Mao, and W.~Zhang.
\newblock How to bridge the gap between modalities: A comprehensive survey on multimodal large language model, 2023.

\bibitem[Tang et~al.(2020)Tang, Niu, Huang, Shi, and Zhang]{scenegraph1}
K.~Tang, Y.~Niu, J.~Huang, J.~Shi, and H.~Zhang.
\newblock Unbiased scene graph generation from biased training, 2020.

\bibitem[Xie et~al.(2023)Xie, Zhao, Xiao, Chan, Li, Xu, Zhang, and Hou]{xie2023dreaminpainter}
S.~Xie, Y.~Zhao, Z.~Xiao, K.~C.~K. Chan, Y.~Li, Y.~Xu, K.~Zhang, and T.~Hou.
\newblock Dreaminpainter: Text-guided subject-driven image inpainting with diffusion models, 2023.

\bibitem[Yu et~al.(2022)Yu, Xu, Koh, Luong, Baid, Wang, Vasudevan, Ku, Yang, Ayan, et~al.]{yu2022scaling}
J.~Yu, Y.~Xu, J.~Y. Koh, T.~Luong, G.~Baid, Z.~Wang, V.~Vasudevan, A.~Ku, Y.~Yang, B.~K. Ayan, et~al.
\newblock Scaling autoregressive models for content-rich text-to-image generation.
\newblock \emph{arXiv preprint arXiv:2206.10789}, 2\penalty0 (3):\penalty0 5, 2022.

\bibitem[Yu et~al.(2023)Yu, Feng, Feng, Liu, Jin, Zeng, and Chen]{yu2023inpaint}
T.~Yu, R.~Feng, R.~Feng, J.~Liu, X.~Jin, W.~Zeng, and Z.~Chen.
\newblock Inpaint anything: Segment anything meets image inpainting.
\newblock \emph{arXiv preprint arXiv:2304.06790}, 2023.

\bibitem[Zhang et~al.(2023{\natexlab{a}})Zhang, Wang, Xu, Wang, and Shi]{zhang2023forget}
E.~Zhang, K.~Wang, X.~Xu, Z.~Wang, and H.~Shi.
\newblock Forget-me-not: Learning to forget in text-to-image diffusion models.
\newblock \emph{arXiv preprint arXiv:2303.17591}, 2023{\natexlab{a}}.

\bibitem[Zhang et~al.(2024)Zhang, Mo, Chen, Sun, and Su]{zhang2024magicbrush}
K.~Zhang, L.~Mo, W.~Chen, H.~Sun, and Y.~Su.
\newblock Magicbrush: A manually annotated dataset for instruction-guided image editing, 2024.

\bibitem[Zhang et~al.(2023{\natexlab{b}})Zhang, Yang, Feng, Qin, Chen, Yu, Chen, Wang, Savarese, Ermon, et~al.]{zhang2023hive}
S.~Zhang, X.~Yang, Y.~Feng, C.~Qin, C.-C. Chen, N.~Yu, Z.~Chen, H.~Wang, S.~Savarese, S.~Ermon, et~al.
\newblock Hive: Harnessing human feedback for instructional visual editing.
\newblock \emph{arXiv preprint arXiv:2303.09618}, 2023{\natexlab{b}}.

\bibitem[Zhu et~al.(2022)Zhu, Zhang, Jiang, Dang, Hou, Shen, Feng, Zhao, Miao, Shah, et~al.]{scenegraph2}
G.~Zhu, L.~Zhang, Y.~Jiang, Y.~Dang, H.~Hou, P.~Shen, M.~Feng, X.~Zhao, Q.~Miao, S.~A.~A. Shah, et~al.
\newblock Scene graph generation: A comprehensive survey.
\newblock \emph{arXiv preprint arXiv:2201.00443}, 2022.

\end{thebibliography}

\clearpage
\appendix

\section{More visual results}
\label{app-visual}

\begin{figure*}[ht]
\centering
\includegraphics[width=0.9\textwidth]{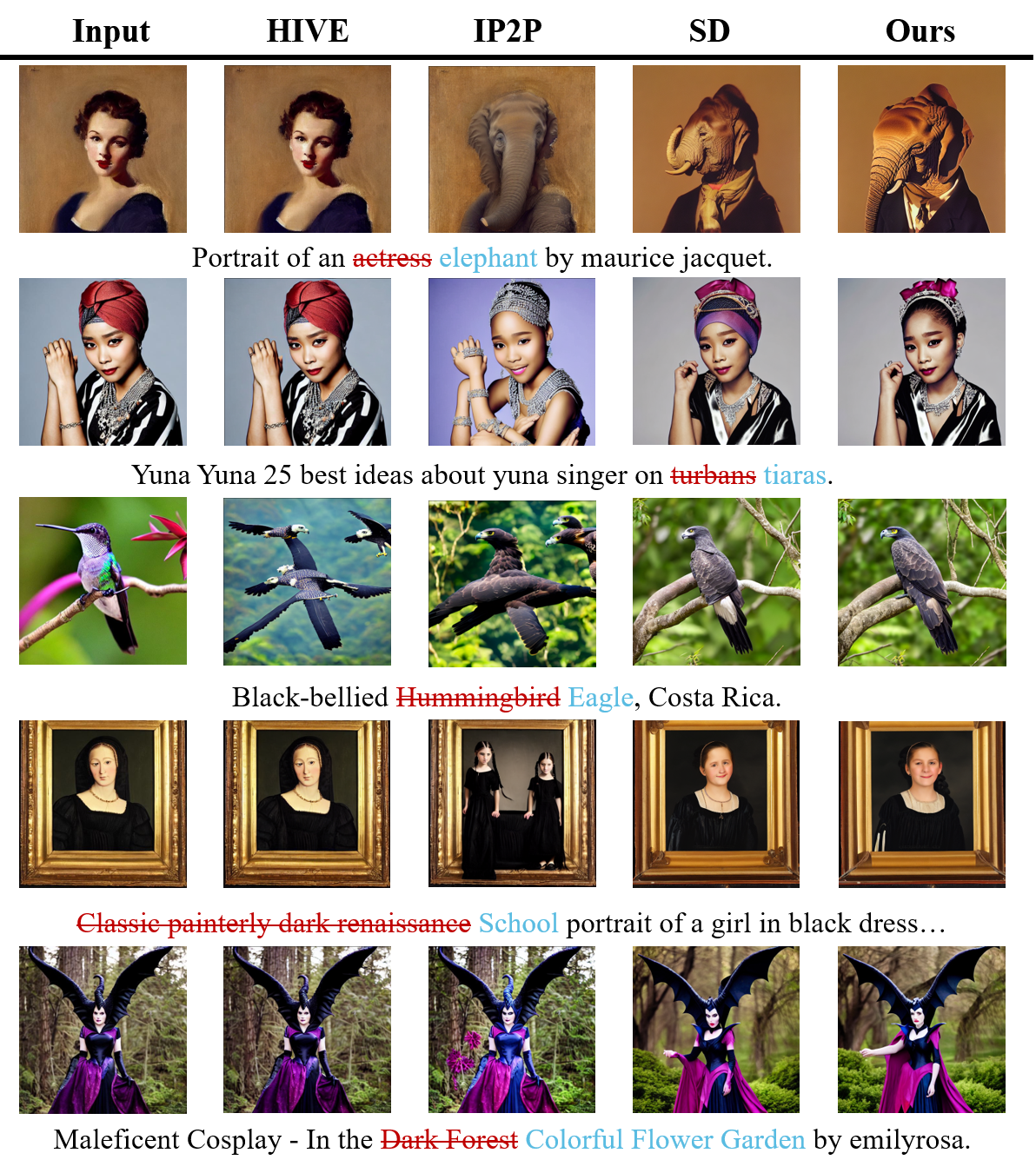}
\caption{Visual comparisons between Hive, IP2P, SD and our method in dataset Human Prompt Generated Dataset. The red annotations indicate the visual concepts in the original image that need to be edited, while the blue annotations represent the new visual elements that should be introduced based on the provided textual prompt.}
\label{fig:app-exp1}
\vspace{-7pt}
\end{figure*}

Figure \ref{fig:app-exp1} demonstrates the results of our proposed method and other editing approaches on the synthetic image dataset mainly about Aesthetics, Human Prompt Generated Dataset. Our results show that our proposed method is highly effective on this aesthetics-centric dataset. The key strength of our approach is that it can make edits aligned not just with local, isolated concepts, but also with the overall style and environmental context of the image. 

\begin{figure*}[ht]
\centering
\includegraphics[width=0.9\textwidth]{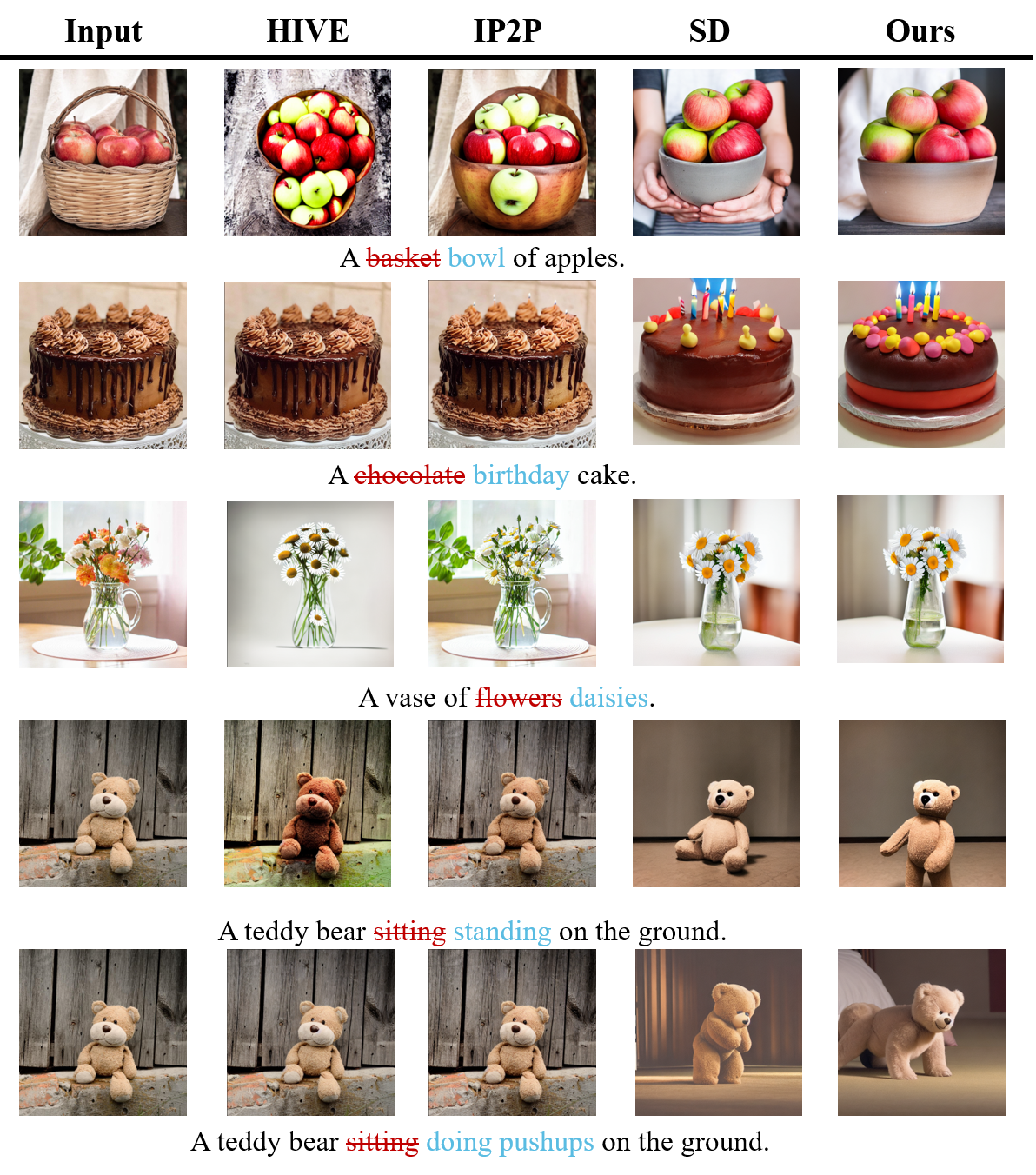}
\caption{Visual comparisons between Hive, IP2P, SD and our method in dataset TedBench. The red annotations indicate the visual concepts in the original image that need to be edited, while the blue annotations represent the new visual elements that should be introduced based on the provided textual prompt.}
\label{fig:app-exp2}
\vspace{-7pt}
\end{figure*}

Figure \ref{fig:app-exp2} demonstrates the results of our proposed method and other editing approaches on the real-world image dataset, TedBench. As can be seen, our method not only performs well on the synthetic images but also exhibits strong performance on the real-world images from the TedBench dataset. 

\section{Metrics}

\subsection{CLIP-T}

The CLIP-T metric is designed to measure the alignment between the textual conditioning (T) and the generated images (I). This metric provides a quantitative assessment of how well the output images match the semantic information conveyed in the input text prompts. The CLIP (Contrastive Language-Image Pretraining) Score serves as a prominent evaluation metric utilized for the assessment and comparison of semantic similarity between images and text in the context of generative models. By evaluating this alignment, we can gain insights into the effectiveness of the text-guided image generation or editing models in capturing and translating the intended visual semantics from the given textual descriptions. Higher scores indicate a greater degree of semantic relevance between the image and text, while lower scores suggest diminished semantic coherence. Specifically, our CLIP-T denotes a metric about alignment between textual conditioning ($t$) and generated images ($i$). 

\begin{align}
     \text{CLIP-T(t,i)} &= \cos(E_{text}(t), E_{image}(i)) \\ &= \frac{E_{text}(t) \cdot E_{image}(i)}{\|E_{text}(t)\| \cdot \|E_{image}(i)\|}, \nonumber
\end{align}

where $cos(\dot)$ denotes the cosine similarity between the text embedding $E_{text}(t)$ and the image embedding $E_{image}(i)$, $E_{text}(t)$ is the text embedding, which is the output of the CLIP text encoder applied to the input text prompt $t$, capturing the semantic meaning and features of the text and $E_{image}(i)$ is the image embedding, which is the output of the CLIP image encoder applied to the generated image I, capturing the visual features and semantics of the image.

\subsection{Inception Score}

The Inception Score (IS) is a widely used metric for evaluating the quality of generated images. It leverages a pre-trained Inception classification model to assess the generated images. The key idea behind IS is that high-quality generated images should have two properties:
1) The model should be highly confident in classifying the objects present in the image (high KL divergence between the conditional label distribution p(y|x) and the marginal label distribution p(y)).
2) The generated images should exhibit diversity, covering a wide range of different object classes (high entropy of the marginal label distribution p(y)).

Mathematically, the Inception Score is calculated as:

\begin{align}
    IS = exp(E_x[KL(p(y|x) || p(y))]),
\end{align}

Where $E_x[\cdot]$ denotes the expectation over the generated images x, and $KL(\cdot \|\cdot)$ is the Kullback-Leibler divergence between the conditional label distribution and the marginal label distribution.

\subsection{L1 Loss}

The L1 loss, also known as the Mean Absolute Error (MAE), is a common metric used to quantify the distribution difference between the original image and the generated image. The L1 loss is calculated as the average of the absolute differences between the corresponding pixel values of the two images. The L1 loss provides a robust and interpretable measure of the overall difference in pixel intensities between the original and generated images. A lower L1 loss indicates that the generated image has a distribution that is more similar to the original image, as there are smaller absolute differences between the corresponding pixel values

Mathematically, the L1 loss between the original image X and the generated image Y is defined as:

\begin{align}
    L1 = E[|X_i - X_o|]
\end{align}

Where $E[\cdot]$ denotes the expected value (average) over all the pixels in the images, $X_i$ denotes input images and $X_o$ denotes output results.

\subsection{CLIP-D}
\label{clip-d}

Despite the high pixel-level similarity between the generated images and the original images, a disconnect in semantic cohesion appears to emerge. This suggests that the visual fidelity of the generated images may not necessarily translate to the preservation of their underlying semantic meaning.  To assess whether the generated images have undergone meaningful editing, rather than merely being copies of the original inputs, we measure the difference in semantic distance between the generated images and their original input and edited versions respectively. Note $X_i$ is the input image, $X_o$ is the output generation and $X_r$ is the revised images as ground truth. We can calculate CLIP-D as follows:

\begin{align}
    CLIP-D(X_o)= (CLIP(X_o,X_i)-CLIP(X_o,X_i))/CLIP(X_o,X_i),
\end{align}
where the $CLIP(\dot)$ denotes the CLIP score between two images. It will allow us to assess whether the editing process has been effective in editing the images, rather than simply creating copies of the original inputs.

\section{Human Study}
\label{app-human}

To comprehensively evaluate the performance of the text-guided image editing methods, we conducted a user study involving over 100 participants from diverse academic backgrounds. Each participant was presented with a series of generated images paired with corresponding textual descriptions. They were asked to rate the images on a scale of 1 to 5 across the following four aspects:
\begin{enumerate}
    \item Alignment: The degree of match between the content and composition of the image, and the semantics and intent expressed in the prompt text.
    \item Fidelity: The level of realism, detail, and overall visual quality of the generated image.
    \item Consistency: The stability and reliability of the text-guided image editing method beween input images and output images.
    \item Preferable: The overall user preference and satisfaction with the generated images, indicating their willingness to use the text-guided image editing technique for their own creative and practical applications.
\end{enumerate}

For each task, we presented the users with a single image along with the same input conditions for every method in each dataset. This approach ensured a consistent and fair evaluation across the different techniques. To mitigate potential biases stemming from preconceived notions of AI-generated images, we employed a data filtering process to exclude low-scoring samples.

After the initial evaluation, we computed the average performance of each method across the diverse datasets. This comprehensive analysis allowed us to derive a holistic understanding of the strengths and limitations of the various text-guided image editing approaches.

Furthermore, we conducted a short survey to gather feedback from the evaluators. By understanding the reasons behind the scoring provided by the participants, we gained valuable insights into the perceived successes and shortcomings of the techniques, which can inform future improvements and development in this field.

\section{Broader Impacts}
\label{app-bi}

The advancements in text-guided image-to-image editing hold immense potential for transforming various industries and enhancing human experiences. As the research community makes significant progress in developing robust models, the applications of this technology extend far beyond the immediate technical challenges.

One of the most compelling areas of impact is the democratisation of visual arts and content creation. By enabling users to easily manipulate and modify images through natural language prompts, the barriers to entry for creative expression are lowered, empowering a wider range of individuals to participate in the creative process. This could lead to a surge in personalised and unique visual content, from marketing materials to educational resources, as hobbyists, small businesses, and those with limited design expertise gain access to powerful image editing tools.

Moreover, the implications of text-guided image editing extend to various industries, such as e-commerce and architecture. Customers could now envision and create custom product designs, while professionals in design-related fields could rapidly iterate on concepts and present photorealistic renderings to clients, accelerating the design process and improving client engagement.

Beyond commercial applications, this technology holds the potential to transform educational and accessibility outcomes. Teachers and students could collaborate to generate illustrative images, enhancing the learning experience, while individuals with visual impairments could better understand and interact with the visual world through text-guided image editing capabilities.

\section{Limitation}
\label{app-limitation}
\begin{figure*}[ht]
\centering
\includegraphics[width=0.9\textwidth]{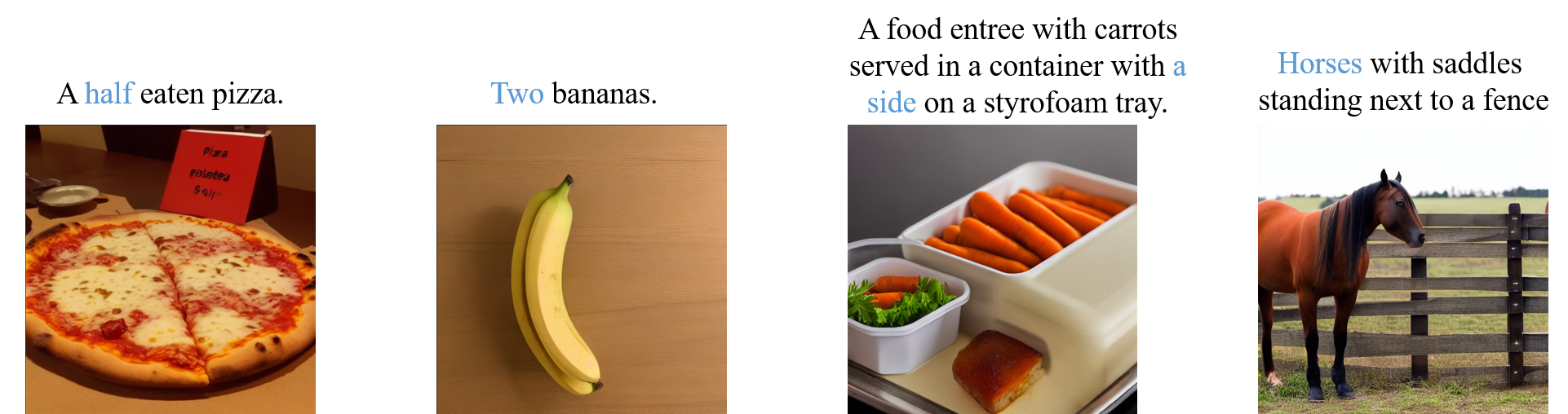}
\caption{Failure visual results in our LaF.}
\label{fig:app=fail}
\end{figure*}

In addition to the strengths highlighted by our performance in experiments, we have also identified several limitations in our text-guided image editing framework, LaF, needs further investigation. Like other multi-modal editing methods, one key challenge limitation we have encountered is the difficulty in precisely quantifying certain editing tasks, such as those involving numerical attributes. While our model has demonstrated strong capabilities in generating visually coherent edits, maintaining accurate control over numeric properties, like object counts or sizes, still remains an area that requires further refinement, as Figure \ref{fig:app=fail} shows.

\end{document}